\documentclass[12pt,journal,compsoc,onecolumn]{IEEEtran}

\usepackage{xcolor} %
\usepackage[T1]{fontenc}
\usepackage{amssymb}
\usepackage{amsmath}

\usepackage{pifont} 
\usepackage{array}  
\usepackage{multirow}
\usepackage{tabularx}
\usepackage{booktabs}
\newcolumntype{M}[1]{>{\centering\arraybackslash}p{#1}}
\usepackage{graphicx}

\def\x{{\bf x}}
\def\S{{\cal  S}}
\def\I{{\cal  I}}
\def\K{{\cal  K}}
\def\Phib{{\bf  \Phi}}

\usepackage[ruled,vlined]{algorithm2e}

\title{Multi-label Classification using Deep Multi-order Context-aware Kernel Networks\thanks{This work is supported by grants from the National Natural Science Foundation of China (No.~62272422, U22B2051).}}
\author{Mingyuan Jiu  $^{\scriptsize (1,2,3)}$ \and
Hailong Zhu $^{\scriptsize (1)}$ \and
Hichem Sahbi $^{\scriptsize (4)}$
\vspace{1cm}
\\ 
{\small (1) School of Computer and Artificial Intelligence, Zhengzhou University, Zhengzhou, China \\ \and
(2) Engineering Research Center of Intelligent Swarm Systems, Ministry of Education, Zhengzhou University, Zhengzhou, China\\ \and
(3) National Supercomputing Center in Zhengzhou, Zhengzhou, China\\ \and
(4) Sorbonne University, CNRS, LIP6, F-75005, Paris, France}}

\begin{document}
 \maketitle
\begin{abstract}

Multi-label classification is a challenging task in pattern recognition. Many deep learning methods have been proposed and largely enhanced classification performance. However, most of the existing sophisticated methods ignore  context in the models’ learning process. Since context may provide additional cues to the learned models, it may significantly boost classification performances. In this work, we make full use of context information (namely geometrical structure of images) in order to learn better context-aware similarities (a.k.a. kernels) between images. We reformulate context-aware kernel design as a feed-forward network that outputs explicit kernel mapping features. Our obtained context-aware kernel network further leverages multiple orders of patch neighbors within different distances, resulting into a more discriminating Deep Multi-order Context-aware Kernel Network (DMCKN) for multi-label classification. We evaluate the proposed method on the challenging Corel5K and NUS-WIDE benchmarks, and empirical results show that our method obtains competitive performances against the related state-of-the-art, and both quantitative and qualitative performances corroborate its effectiveness and superiority for multi-label image classification. \\

\noindent {\bf keywords.} {Multi-label classification \and Context-aware kernel \and Deep learning \and Deep unfolding.}
\end{abstract}

\section{Introduction}
 Multi-label image classification is a challenging task in pattern recognition. It aims at  identifying  the presence of objects, scenes, or concepts  by assigning multiple labels to images.  This task is crucial for parsing and understanding visual information, significantly enhancing machine cognition of complex visual scenes. Multi-label classification can also be applied to many scenarios, such as human attribute recognition, scene understanding, image tagging, labeling, and so on. However, multi-label classification encounters many challenges~\cite{alazaidah2016trending}, primarily due to the complexity and diversity of the image contents. Compared to single-label classification~\cite{deng2009imagenet}, this task requires the model to simultaneously recognize all relevant objects and concepts in an image, and annotate them accurately, demanding higher requisite on the model’s discrimination and generalization ability. The relationships among objects in an image can also be exceedingly complex, including, but not limited to, exclusivity, dependency, and hierarchical relationships. Additionally, long tail  label distributions  further increases the difficulty of multi-label classification.
 
 Recently, the rapid development of deep learning technologiess~\cite{wang2016cnn}, especially
the introduction of transformers~\cite{guo2019visual} and attention mechanisms~\cite{vaswani2017attention}, along with label relationship learning through Graph Convolutional Networks (GCN)~\cite{chen2019multi,sahbi2021learning}, has significantly advanced multi-label classification performance. These edge-cutting methods, that learn intricate dependencies between pixels, regions, or labels, have significantly improved  recognition and classification performances in complex scenes. Despite these advancements, challenges remain in order to fully leverage contextual information and structural relationships among objects within images.

It is well known that appropriately leveraging contextual information into a learning model can enhance performances~\cite{tamura2021qpic,sahbi2010context}. Following this line, this work proposes a novel multi-label classification framework that learns rich contextual information within images through structure-aware perception. Based on a deep understanding of the importance of context and using multi-layer deep networks, our framework effectively captures complex and fine-grained relationships in images.  This is achieved by learning these complex relationships as a part of a carefully designed kernel function. The latter allows obtaining a significant gain in accuracy and robustness of multi-label classification. Considering the aforementioned issues, the main contributions of this work include

\begin{itemize}
	\item A novel multi-label classification framework that combines contextual information through a multi-order context-aware kernel network (MCKN), resulting in more discriminative features;
	\item An end-to-end framework that learns the geometrical relationships between image regions with increasing contextual ranges;
	\item And extensive experiments on several benchmarks which show that our method obtains very competitive results and significantly outperforms different baselines as well as the related  work.
\end{itemize}

\section{Related Work}

\subsection{Multi-label classification}
The study of multi-label classification has attracted increasing attention in recent years. Initial efforts primarily focused on generating region proposals through object detection techniques for label prediction~\cite{wei2015hcp}. Subsequent region-based work  delved into modeling spatial dependencies among objects. For instance, Wang et al.~\cite{wang2017multi} proposed a model utilizing spatial transformer layers and Long Short-Term Memory (LSTM) units in order to capture spatial dependencies between different object areas in images. Chen et al.~\cite{chen2018recurrent} explored semantic interactions between labels by leveraging label co-occurrence. Wu et al.~\cite{wu2021gm} used graph-matching techniques to simultaneously explore spatial associations between instances, semantic dependencies of labels, and the feasibility of instance-label matching.

Recently, many studies have dedicated efforts to capturing relationships between labels. Sequence-based methods analyze and learn semantic associations between label vectors by using Recurrent Neural Networks (RNN), while graph-based approaches capture and utilize label dependencies through Graph Convolutional Networks (GCN). For example, Chen et al.~\cite{chen2019multi} mapped complex label relationship graphs into series of independent label classifiers. Moreover, Wang et al.~\cite{wang2020multi} constructed label graphs by analyzing label co-occurrence information in the data for label representation learning.

The emergence of vision transformers~\cite{dosovitskiy2020image} has introduced a new direction for
multi-label classification. Lanchantin et al.~\cite{lanchantin2021general} developed a framework based on transformer encoders in order to capture complex dependencies between visual features and labels, while Liu et al.~\cite{liu2021query2label} explored the use of label embeddings to directly query the presence of labels in images using transformer decoders.

Although the aforementioned multi-label classification methods have made significant progress, particularly in considering spatial dependencies and interactions between labels, there is, however, still a lack of in-depth utilization of structural and contextual information within images. Therefore, our work focuses on further exploring these aspects, aiming to capture rich contextual information by learning and leveraging geometric relationships in images at multiple orders and ranges. Our proposed approach is intended to empower the multi-label classification model with more discriminative image representations.

\subsection{Context-aware models}
The concept of "Context Awareness" has been extensively studied across multiple fields, particularly in computer vision \cite{li2011superpixel,sahbi2010context,jiu2019deep}, where its applications span a wide extent of applications including object detection and recognition, scene understanding, image segmentation, multi-label classification, etc. Early work on context information primarily focused on integrating local features within images with their surrounding contextual information for object recognition and scene classification. Since the advancement of machine learning techniques, such as random forests and support vector machines (SVM), context-dependent scene modeling has been a major bottleneck in enhancing performances in different classification tasks. For instance, Torralba et al.~\cite{oliva2006building} explored methods to improve object detection accuracy through the use of scene context.

With the advancement of deep learning, context modeling in computer vision has led to a major breakthrough. For instance, the VGG network~\cite{simonyan2014very} captures multi-level features of images through convolutional layers, while the Faster R-CNN~\cite{ren2015faster} utilizes a region proposal network to precisely focus on key parts of images, enhancing detection performance. Furthermore, Graph Convolutional Networks~\cite{scarselli2008graph,sahbi2021learning,mazari2019mlgcn} and Vision Transformers (ViT)~\cite{dosovitskiy2020image} enhance the processing of context information by capturing relationships between nodes in graphs and by employing global self-attention mechanisms, respectively. Additionally, multi-modal context modeling with the Bilinear Attention Networks (BAN)~\cite{kim2018bilinear} offers rich scene knowledge for other tasks, namely visual question-answering.

In this paper, we model and learn contextual relationships between image regions at multiple ranges and orders. This modeling leads to better image representations and to conceptually a different approach compared to the related work; this approach is based on learning multiple order similarity kernels whose underlying unfolded networks allow capturing both content and geometric structure (context) in the learned image representations. Our structural relationships in the unfolded networks are dynamically learned end-to-end.

\begin{figure}[tb]
	\centering
	\includegraphics[width=0.9\linewidth]{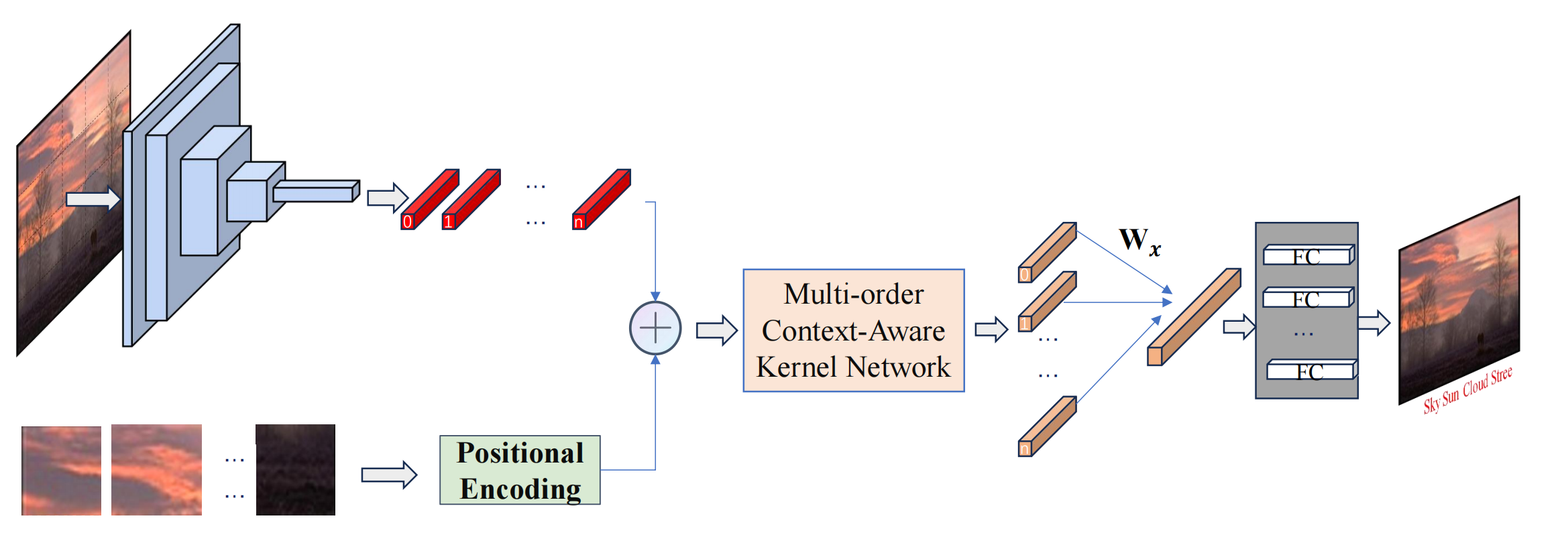}
	\caption{Deep Multi-order Context-aware Kernel Network framework.} \label{fig:architecture}
	\vspace{-8pt}
\end{figure}
\begin{figure}
	\centering
	\includegraphics[width=0.6\linewidth]{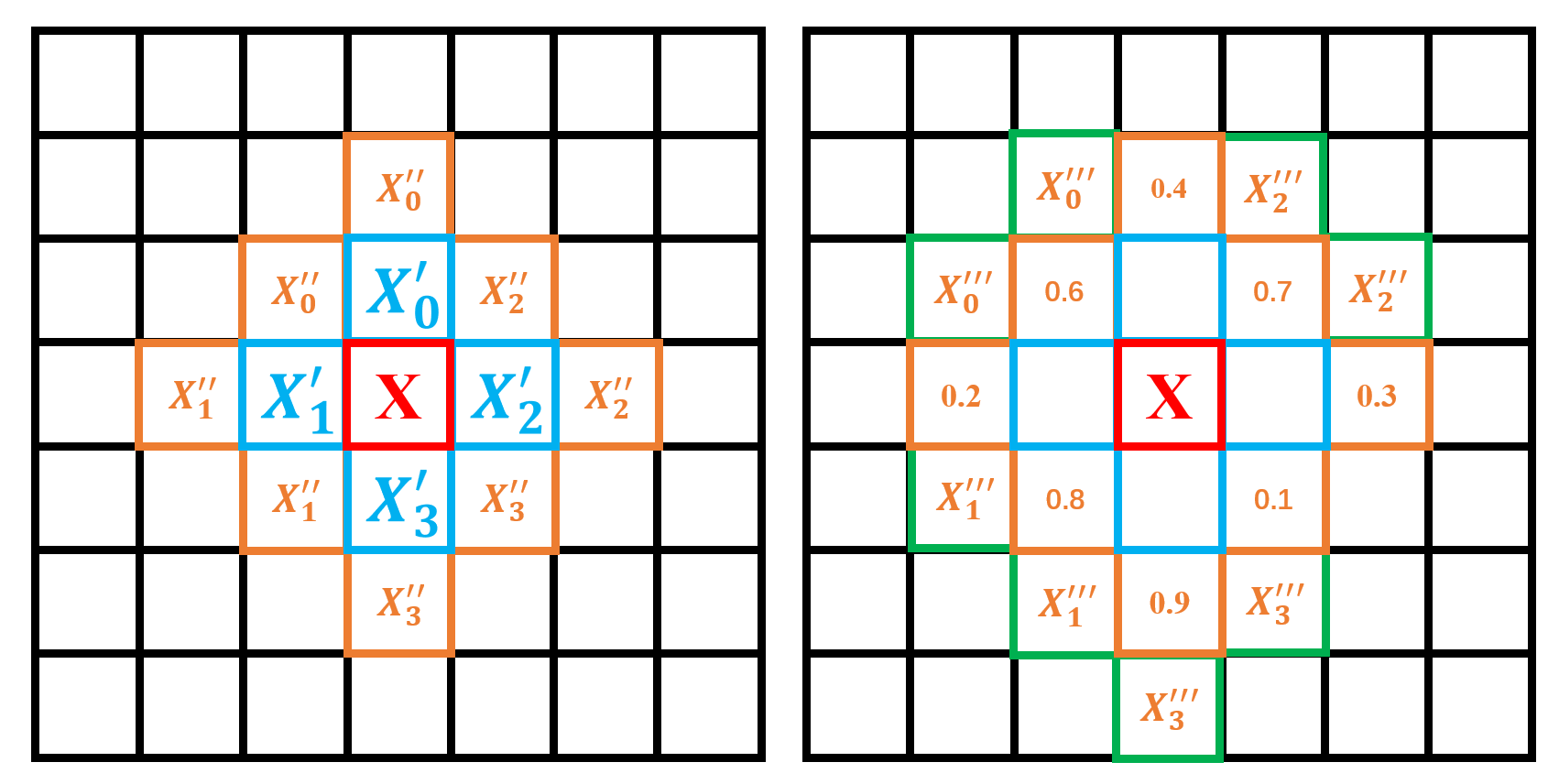} 
	\caption{Multi-order neighborhood system. The left side shows the first-order and second-order neighborhoods. On the right, the third-order neighborhood is built from the second-order neighborhood based on the transition probabilities.} \label{fig:neighbors}
\end{figure}
\section{Method}
In this section, we consider  a multi-order neighborhood system that allows  integrating contextual information of image regions,  and also defining rich and more discriminative image representations. A given image is segmented into a regular grid of cells, with each cell being described with (i) visual  features obtained through a pretrained   model, and (ii) positional features obtained by encoding their location  in  images.  For each cell,  the integrated features are then fed to our proposed multi-order context-aware kernel network, which updates cell features by incorporating their first and higher-order neighbors. The overall structure of the network is shown in Fig.~\ref{fig:architecture}.

\subsection{Context-aware Kernel Map}
For simplicity, we define  $\{\I_p\}_{p=1}^P$ as a set of labeled training images, $Y_k^p$ is a binary variable standing for the membership of a given image $\I_p$ to the class $k \in \{1, \ldots, K\}$. $\S_p = \{\x_1^p, \ldots, \x_n^p\}$  corresponds to a set of non-overlapping cells extracted from a regular grid in $\I_p$; without a loss of generality, $n$ is constant for all images. 

The similarity between any two images $\I_p$ and $\I_q$ can be measured by using a convolution kernel:

\begin{equation}
	\K(\I_p, \I_q) = \sum_{i, j} \kappa(\x_i^p, \x_j^q), 
	\label{eq:1}
\end{equation} 
\def\X{{\cal X}}
\def\KK{{\bf K}}
\def\SB{{\bf S}}
\def\P{{\bf P}}

\noindent here $\kappa$ is a positive definite elementary kernel, such as linear, polynomial, and Gaussian,  or their linear combinations. These elementary kernels primarily focus on the visual content of the cells within images, ignoring their contextual relationships.

In order to obtain a more relevant similarity,  we define a learned context-aware kernel $\kappa$ (or equivalently its Gram matrix,  denoted as $\KK$, where $[\KK]_{\x_i,\x_j} = \kappa(\x_i,\x_j)$ and $\x_i,\x_j \in \X$ among all cells $\X = \bigcup_p \S_p$). The kernel matrix $\KK$ is obtained as   \cite{jiu2014human,sahbi2015imageclef}
\begin{equation}
	\min_\KK tr(-\KK \SB^\top) - \alpha \sum_{c=1}^C  tr(\KK \P_c \KK^\top \P_c^\top) + \frac{\beta}{2} \| \KK \|_2^2 ,
	\label{eq:2}
\end{equation}
\noindent here $\alpha \geq 0$, $\beta > 0$,  $\SB$ is the similarity matrix of data in $\X$ without context information,  $\top$ denotes matrix transpose, and $tr$ denotes the trace operator. The  set of matrices  $\{\P_c\}_{c=1}^C$ defines  the neighborhood relationships between cells (in practice $C=4$,  corresponding to the four directions: up,  down,  left and right). Specifically, for a given cell $\x$, if there exists an immediate  neighbor $\x'$ in direction $c$, then $[\P_{c}]_{\x,\x'} \neq 0$; otherwise $[\P_{c}]_{\x,\x'} = 0$,  $\forall \x' \in \X$.  In Eq.~\eqref{eq:2}, the leftmost part is a fidelity criterion that provides high kernel values for visually similar cells pairs $\{(\x_i, \x_j)\}_{ij}$, the second term strengthens or weakens the kernel values between these pairs based on the similarity of their neighborhood, and the right-hand side term acts as a regularizer controlling the smoothness of the learned kernel solution.
\begin{figure}[tbp]
	\centering
	\includegraphics[width=0.9\linewidth]{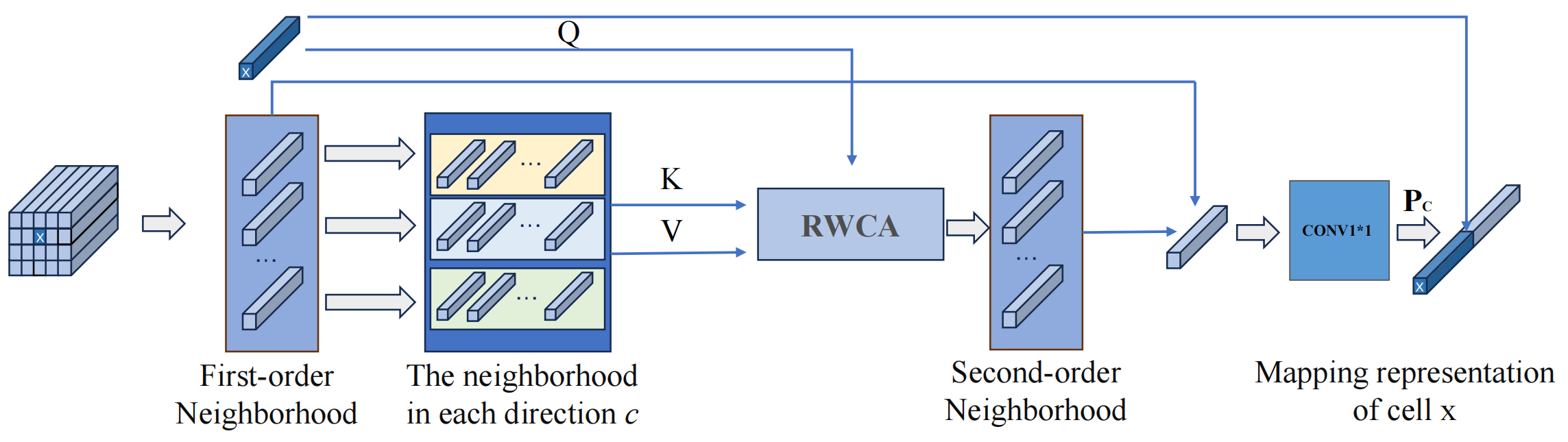} 
	\caption{Details of the Deep Multi-order Context-aware Kernel Network.  ``RWCA'' is  the abbreviation of Random Walk and Context Awareness.} \label{fig:details}
	\vspace{-8pt}
\end{figure}

One may show that the minimization of Eq.~\eqref{eq:2} leads to the following recursive solution:
\begin{equation}
	\KK^{(t+1)} = \SB + \gamma \sum_{c=1}^C \P_c \ \KK^{(t)} \ \P_c^{\hspace{-0.05cm}\top},
	\label{eq:3}
\end{equation}
\noindent where $\gamma = \frac{\alpha}{\beta}$ controls the impact of context, guaranteeing that learning converges  to a stable solution. Eq.~\eqref{eq:3} can be written  using an explicit kernel mapping form as
\begin{equation}
	\Phib^{(t+1)} = \left( {\Phib^{(0)}}^{\hspace{-0.1cm}\top} \  \gamma^{1/2} \P_1 {\Phib^{(t)}}^{\hspace{-0.1cm}\top}  \ldots \  \gamma^{1/2} \P_c {\Phib^{(t)}}^{\hspace{-0.1cm}\top}  \right)^{\hspace{-0.1cm}\top},
	\label{eq:4}
\end{equation}
where the matrix $\Phib^{(0)}$ represents either an exact or an approximate kernel mapping of $\SB$, and the matrix $\Phib$ is the kernel mapping of $\KK$ that needs to be estimated. This update procedure can be reformulated using a fixed  feed-forward network structure;  {each layer in this network corresponds to an iteration: the input layer $\Phib^{(0)}$ is extracted through a pre-trained visual model (such as ResNet101, TresnetL,  etc.),  and the intermediate hidden layers $\{\Phib^{(t)}\}_t$ are updated through the neighborhood matrices $\{\P_c\}_c$ and the output of the previous layers according to Eq.~\eqref{eq:3},  whilst  the final output layer (denoted as $\phi_{\cal K}({\cal S}_p)$) corresponds to  the high-order context-aware features of the image $\I_p$ which are generated by iteratively aggregating multi-order contextual information as described  subsequently.}

\subsection{High-order Context}
In the previous section,  we discussed how the  first-order neighborhood is leveraged in kernel-based representation learning.  For  scenes with short range contextual relationships (particularly when individual small objects are considered),  first-order neighborhoods are enough. However, for  scenes with wider range contextual relationships (for instance when mutiple objects co-exist), first-order neighborhoods are not sufficient. As suggested by~\cite{jiu2022neuro}, wider range contexts are involved by extending the scope of first-order neighborhood but it may also significantly increases the dimensionality of the learned representations,  and hence the computational cost,  and may also potentially introduce excessive noise. Therefore,  our proposed contribution relies on higher-order neighborhoods, using random walk and self-attention mechanisms that aggregates more relevant context from a wider range, thus avoiding excessive noise and reducing unnecessary computational burden.

{Random walk is able to expand  low-order neighborhoods to high-order ones,  effectively filtering out  noisy image areas.  In other words,  by moving stochastically through cell neighbors,  transitions  can be adjusted according to contextual information  to balance the exploration of  neighboring cells  and the exploitation of local ones.  This approach prevents getting stuck in noisy cells,  and it allows capturing  more meaningful contextual information.  Multiple independent iterations of random walk tend to retain cells  that are frequently visited and more relevant to the content of the central cell,  gradually filtering  out noisy image areas  and enhancing the overall quality of the learned representations.}

Formally,   given a cell $\x$, we define its first-order (c-typed) neighborhood $\{{\cal N}_c^{(1)}(\x)\}_c$ through
the set of matrices $ \{\P_c \}_{c=1}^C$  where values of $c$ refer to different types of neighborhoods.  For any order $p\geq 2$,  the p-th (higher) order  neighborhood of $\x$  is recursively defined  as 
\begin{equation}
{\cal N}_c^{(p)}(\x)=\displaystyle \bigcup_{\x'\in {\cal N}_c^{(1)}(\x)} {\cal N}_c^{(p-1)}(\x') \ \ \  \ \ \  \textrm{with} \  \x' \neq \x.
\end{equation} 

The attention score between $\x$ and any $\x''$ in  ${\cal N}_c^{(p)}(\x)$  is obtained  as 
\begin{equation}
	\text{score}(\phi(\x), \phi(\x'')) = \text{softmax}\left(\frac{W_q{\phi(\x)}(W_k \phi(\x''))^{\hspace{-0.05cm}\top}}{\sqrt{d}}\right),
	\label{eq:5}
\end{equation}
\noindent where $\phi(\x)$ refers to   the features of the target  cell $\x$,  and $\phi(\x'')$ denotes the features of a cell $\x''$ within the neighborhood ${\cal N}_c^{(p)}(\x)$.  $W_q$ and $W_k$ are learnable parameter matrices and $d$ corresponds to the dimensionality of the keys.

In order to consider the transition probabilities from the first order neighborhood to the second-order one,  we evaluate  the probability $p_c^{(p)}(\x, \x'')$ (with $p=2$) by employing an exponential function to the attention scores and then normalize the values by summing all the scores in the second-order neighborhood ${\cal N}_c^{(2)}(\x)$  for the target cell $\x$ and the direction $c$ so that  their sum equates 1,  i.e., 

\begin{equation}
  p_c^{(2)}(\x, \x'') = \frac{\exp(\text{score}(\phi(\x), \phi(\x'')))}{\sum_{z\in {\cal N}_c^{(2)}(\x)} \exp(\text{score}(\phi(\x), \phi(z)))}. 
  \label{eq:6}
\end{equation}
\noindent Here  $p_c^{(2)}(\x, \x'')$ defines  the probability of a random walk from the first-order to the second-order context for cell $\x$ in direction $c$.

Subsequently, using the aforementioned transition probabilities, we obtain a better reestimate of the features while taking into account the second-order neighborhood as 
\begin{equation}
	\phi_{c,p}(\x) = \sum_{{ \x''\in {\cal N}_c^{(p)}(\x) }} p_c^{(p)}(\x, \x'') (W_v\phi(\x'')), \ \ \ \textrm{with} \ p=2, 
	\label{eq:7}
\end{equation}
\noindent where $W_v$ is a learnable parameter matrix used to transform the features; it is  multiplied by the transition probabilities to form the second-order (c-typed) contextual features $\phi_{c,2}(.)$.

In order to build higher-order contexts,  we employ an iterative method similar to that used for the second-order context, as shown in Fig.~\ref{fig:neighbors}.  However,  it is important to note that higher-order contexts involve information with larger  distances,  and the relevance between a given central cell and the cells in the higher-order neighborhood decreases. Therefore,  instead of using all first-order neighboring cells to construct second-order contexts,  only a part of cells is  chosen in order  to build higher-order neighborhoods using transition probabilities.

\subsection{Deep Multi-order Context-aware Kernel Networks}
In this section, we build context-aware kernel mapping using the underlying multi-layered  networks. This process is achieved iteratively. Below, we detail each step of the network’s construction procedure.

In order to capture both  content and context of a given  cell $\x$  at the $t$-th layer,  we define  the multi-order representation as the concatenation  of  all the orders of contextual features through different directions  $c \in \{1,\dots,C\}$ as
\def\Psib{{\bf \Psi}}
\begin{equation}
	\phi^{(t)}_{c}(\x)   = \bigg(\phi_{c,1}^{(t)}(\x)^{\hspace{-0.05cm}\top} \ \   \phi^{(t)}_{c,2}(\x)^{\hspace{-0.05cm}\top}  \ \ \ldots\bigg)^{\hspace{-0.05cm}\top},  
\end{equation}
then the representation  at the $(t+1)$-th layer for a given cell $\x$ is obtained  by integrating all the directions as 
\begin{equation}
	\Phib^{(t+1)}= \bigg({\Phib^{(t)}}^{\hspace{-0.1cm}\top}   \ \  \gamma^{1/2} \P_{1}  {\Phib^{(t)}_{1}}^{\hspace{-0.1cm}\top}  \ \  \ldots \ \ \gamma^{1/2} \P_{C} {\Phib^{(t)}_{C}}^{\hspace{-0.1cm}\top}\bigg)^{\hspace{-0.15cm}\top},
	\label{eq:9}
\end{equation}
\noindent being $\phi^{(t)}_{c}(\x)$ a column of ${\Phib^{(t)}_{c}}$,  and similarily $\phi^{(t)}(\x)$ a column of ${\Phib^{(t)}}$. The details of deep multi-order context-aware kernel network are shown in Fig.~\ref{fig:details}.

Eq.~\eqref{eq:10} details how, at each layer, the kernel value  $[\KK^{(t)}]_{\x_i,\x_j}$  between
two cells $\x_i$ and $\x_j$ is evaluated by unfolding the map of the kernel as  
\begin{equation}
	[\KK^{(t)}]_{\x_i,\x_j} = \phi^{(t)}(\ldots(\phi^{(1)}(\phi^{(0)}(\x_i)))) \cdot \phi^{(t)}(\ldots(\phi^{(1)}(\phi^{(0)}(\x_j)))), 
	\label{eq:10}
\end{equation}
It's worth noticing that  the dimensionality of $\phi(\x)$ increases with deeper networks and  concatenation of multi-order contextual features. To address the resulting computational challenge, we introduce $1\times1$ convolutions at each layer of the context-aware kernel map network for dimensionality reduction. To effectively preserve essential features, we implement a layerwise   dimensionality reduction using~\cite{li2021more} as
\begin{equation}
	\psi^{(t)}_{c}(\x) = C_t ( \phi^{(t)}_{c}(\x)).
	\label{eq:11}
\end{equation}
\noindent Here  $\phi^{(t)}_{c}(\x)$  and $\psi^{(t)}_{c}(\x)$ respectively refer to the contextual features before and after undergoning dimensionality  reduction,  and $C_t(.)$ stands for the convolution operation.

Subsequently,  Eq.~\eqref{eq:12} redefines the similarity between two images $\I_p$ and $\I_q$  
using this kernel construction (following  Eq. \ref{eq:1})
\begin{equation}
	\K(\I_p, \I_q) = \sum_{\x_i \in \S_p} \phi^{(t)}(\ldots(\phi^{(0)}(\x_i))) \cdot \sum_{\x_j \in \S_q} \phi^{(t)}(\ldots(\phi^{(0)}(\x_j))).
	\label{eq:12}
\end{equation}
Eq.~\eqref{eq:12} reveals the inner product between two recursive kernel mappings, with each one corresponding to an unfolded multi-layered neural network whose feature mappings capture broader contexts as the depth of this network increases. The network’s structure is similar to common deep learning architectures, yet distinct in that the network depth and the number of units per layer are dynamically determined based on the dimensions of the kernel mappings and the number of iterations prior to convergence. 

When considering the limit of Eq.~\eqref{eq:3} as $\tilde{\KK}$ and the underlying map in Eq.~\eqref{eq:4} as $\tilde{\phi}(.)$,  the  convolution kernel $\K$ between two given images $\I_p$ and $\I_q$ can be expressed as
\begin{equation}
	\K(\I_p, \I_q) = \langle \tilde{\phi}_{\mathcal{K}}(\S_p),\tilde{\phi}_{\mathcal{K}}(\S_q) \rangle, 
	\label{eq:13}
\end{equation}
\begin{equation}
	\tilde{\phi}_{\mathcal{K}}(\S_p) = \sum_{\x_i  \in \S_p}  w_i \tilde{\phi}(\x_i),
	\label{eq:14}
\end{equation}
\noindent being  $\{w_i\}_i$ learnable parameters.   Hence, each constellation of cells in a given image $\I_p$ can be represented by a deep explicit kernel map $\tilde{\phi}_{\mathcal{K}}(\S_p)$ that aggregates the representation of all the cells in $\I_p$. In order to explore the full potential of Eq.~\eqref{eq:13}, we consider an end-to-end framework that learns the neighborhood system $\{\P_c\}_c$ within images.   
\begin{table}[t]
	\vspace{-10pt}
	\caption{Comparisons (in percentages) of different methods with ours in terms of Recall (R), Precision (P), and F1 Score (F1) on the Corel5K dataset.}
	\label{tab1}
	\centering
	\resizebox{0.9\linewidth}{!}{
	\begin{tabularx}{\textwidth}{c|c|c|c|c|c} 
		\toprule
		\makebox[0.3\textwidth][c] {\textbf{Method}} & 	\makebox[0.2\textwidth][c] {\textbf{Backbone}} & \makebox[0.15\textwidth][c] {\textbf{CL}} &\makebox[0.1\textwidth][c] {\textbf{R}} & \makebox[0.1\textwidth][c] {\textbf{P}} & \makebox[0.1\textwidth][c] {\textbf{F1}}\\
		\midrule
		FT DMN+SVM\cite{jiu2022neuro} & - & no & 38.1 & 23.4 & 28.9 \\
		CNN-R \cite{murthy2015automatic}& - & no & 41.3 & 32.0 & 36.0 \\
		3-layer DKN+SVM \cite{jiu2017nonlinear}& - & no & 43.2 & 25.6 & 32.1 \\ 	\hline
		LNR+2PKNN \cite{zhang2018neural}&- & yes & 46.1 & \textcolor{blue}{44.2} & 44.9 \\
		DCKN\cite{jiu2022neuro} & ResNet101 & yes & 44.4 & 33.4 & 38.1 \\
		Q2L-TResL \cite{liu2021query2label}     & TResNetL&no&48.1& 43.5&45.7\\
		\hline
		\multirow{3}{*}{DMCKN (ours)} & TResNetL & 4*5 & 47.8 & 43.4 & 45.5 \\
		& TResNetL & 8*10 &\textcolor{blue}{ 48.3 }& 43.9 &  \textcolor{blue}{45.9} \\
		& Cvt-w24 & 8*10 & \textcolor{red}{49.1} & \textcolor{red}{45.2} & \textcolor{red}{47.0} \\
		\bottomrule
	\end{tabularx}
}
	\vspace{-10pt}
\end{table}

\begin{table}[tb]
	\caption{Ablation study of the context-aware module and group fully connected layer on Corel5k dataset.}
	\label{tab2}
	\centering
	\begin{tabularx}{0.68\textwidth}{c|c|c|ccc} 
		\toprule
		\makebox[0.2\textwidth][c]{\textbf{Method}} & \makebox[0.08\textwidth][c]{\textbf{CA}} & \makebox[0.08\textwidth][c]{\textbf{LG}} &\makebox[0.08\textwidth][c]{\textbf{R}} & \makebox[0.08\textwidth][c]{\textbf{P}} & \makebox[0.08\textwidth][c]{\textbf{F1}}\\
		\midrule
		Baseline & \ding{55} & \ding{55} & 45.9& 38.3&41.7 \\  \hline
		\multirow{3}{*}{Ours} & \ding{51}  & \ding{55} &47.1 & 40.2&43.3 \\
		& \ding{55} & \ding{51} &46.3 &38.9&42.3 \\
		& \ding{51}  & \ding{51} &\textbf{47.5} & \textbf{40.9}&\textbf{ 43.9}\\
		\bottomrule
	\end{tabularx}
\end{table}

\subsection{End-to-end Supervised Learning}
We train our context-aware kernel map network (end-to-end) and particularly its underlying contextual parameters, for the task of multi-label classification. Considering $N$ training images $\{\I_p\}_{p=1}^N$ and their category labels $Yp^k$, where  $Y_p^k=1$  if $\I_p$  belongs to the $k^\textrm{th}$ category, and $Y_p^k=-1$ otherwise. In our kernel map network,  we use a fully connected layer for classification. To address class imbalance, we consider a grouped training strategy based on label co-occurrence, by training a classification layer for each group and by weighting the underlying losses in order to obtain the total loss; the latter  is defined as 
\begin{equation}
	\min_{{\{W_g\}, \{\P_c\}}} \frac{1}{2} \sum_{g=1}^{G} \|W_g\|_2^2 + \sum_{g=1}^{G} C_g \sum_{p=1}^{N_g} \mathcal{L}_g(W_g \phi_\K(\S_p), Y_{p, g}^k), 
\end{equation}
\noindent here \(W_g\) is the weight matrix for the fully connected layer, $\{\P_c\}_c$ are the learnable context  matrices,  $N_g$ the size of each group,  and \(C_g\) corresponds to a hyper-parameter. The total loss includes the weighted sum of cross-entropy losses $\mathcal{L}_g$ and $\ell_2$ regularization across category groups. Error backpropagation and gradient descent algorithm are used to update the parameters.

\begin{table}[tb]
	\caption{Ablation study of network depth and context-awareness levels: analysis of recall (R), precision (P), and F1 Score (F1) on Corel5k dataset (R/P/F1).}
	\label{tab3}
	\centering
	\begin{tabular}{c|c|c|c} 
		\toprule
		\ & \textbf{One-layer} & \textbf{Two-layers} & \textbf{Three-layers} \\
		\midrule
		SC & 47.1/39.8/43.1 & 47.5/40.9/43.9 & 47.9/41.7/44.5 \\
		TC & 47.4/40.2/43.5 & 47.9/41.3/44.3 & \textbf{48.3}/\textbf{42.2}/\textbf{45.0} \\
		\bottomrule
	\end{tabular}
\end{table}

\begin{table}[tb]
	\caption{\textcolor{black}{Ablation study on the random walk strategy in the Corel5k dataset. $thres$ shows the probability threshold.}}
	\label{tab4}
	\centering
	\begin{tabularx}{0.68\textwidth}{c|c|ccc} 
		\toprule
		\makebox[0.2\textwidth][c]{\textbf{Method}} & \makebox[0.2\textwidth][c]{\textbf{RWS}} &\makebox[0.08\textwidth][c]{\textbf{R}} & \makebox[0.08\textwidth][c]{\textbf{P}} & \makebox[0.08\textwidth][c]{\textbf{F1}}\\
		\midrule
		\multirow{5}{*}{DMCKN(Ours)} & \ding{55} & 46.59 & 39.26 & 42.61 \\  
		& $thres=0$ & 48.11 & 40.98 & 44.26 \\
		& $thres=0.62$ & 48.03 & 41.21 & 44.35 \\
		& $thres=0.67$ & \textbf{47.96} & \textbf{41.32} & \textbf{44.39} \\
		& $thres=0.70$ & 47.91 & 41.33 & 44.37 \\
		\bottomrule
	\end{tabularx}
\end{table}
\section{Experiment}
\subsection{Implementation Details}
We evaluate our framework on the Corel5K and NUS-WIDE benchmarks, which is trained in 200 epochs with an AdamW optimizer, a batch size of 128, and a maximum learning rate of $0.0001$. An early stopping strategy is used, with data augmentation techniques such as RandAugment and Cutout, and exponential moving average applied to model parameters with a decay rate of $0.9997$.

\subsection{Results on Corel5K }
The Corel5k dataset comprises 4999 images annotated with 260 concepts. It  is split into 4500 training and 500 testing images, with each test image potentially labeled with up to 5 keywords. Performance metrics include average precision (P), recall (R), and F1-score (F). Images are resized to $400\times500$ pixels and divided into $4\times5$ and $8\times10$ cell  configurations for analysis. We use Resnet101, TresnetL~\cite{ridnik2021tresnet}, and Cvt~\cite{wu2021cvt}, pre-trained on ImageNet to extract visual features. To address category imbalance, we select all positive instances and three times random subset of negative ones.

Tab.~\ref{tab1} compares the performance of our model (DMCKN) against other models on the Corel5K dataset. ``CL''  stands for  context learning. The best and second-best performances for each metric are highlighted in red and blue, respectively. The results demonstrate that models with context significantly outperform those without context. Under two different scene configurations (4$\times$5 and 8$\times$10), DMCKN  shows superior performance in both configurations. Furthermore, performance  on the latest Cvt network architecture further validate our model's robustness to different backbone networks.

\noindent \textbf{Ablation study.} {We conducted an in-depth ablation studies to assess the impact of various modules on the performance, focusing on five core components: context awareness, group fully connected layers, context-aware distance, network depth, and the random walk strategy}. These experiments adopted ResNet101, pre-trained on the ImageNet dataset, as the feature extractor.

{Firstly, we evaluate the effectiveness of the context awareness module and group fully connected layers by comparing model's performance with and without these modules. Keeping other configurations fixed, we adjuste the context-aware distance (second-order and third-order neighborhoods) to explore the specific impact of different context-aware distances on performance. Additionally, we evaluate the model's performance at different network depths to investigate the impact of network depth on performance. Finally, we also study the role of different random walk strategies in mitigating model noise by using different random walk strategies.}

Tab.~\ref{tab2} shows the results of ablation studies on the context awareness module (\textcolor{red}{CA}) and the group fully connected layer (\textcolor{red}{LG}) in our model. The results show that the context awareness module increases the performance by 1.2/1.9/1.6, and the group fully connected layer obtained performance gain of  0.4/0.6/0.6. When both are used, there is a significant performance enhancement of 1.6/2.6/2.2, validating the effectiveness of our modules. 

Tab.~\ref{tab3} demonstrates the impact of second-order context awareness (SC) and third-order context awareness (TC) on model performance across different network depths (one-layer, two-layers, and three-layers). The results indicate a gradual improvement in model's performance with increasing network depth. This suggests that deeper network structures can lead to better performance improvements when considering more profound levels of context information.

{Tab.~\ref{tab4} presents the impact of different random walk strategies (RWG) on model's performance. Here, \ding{55} indicates that the random walk strategy was not used to construct higher-order neighborhoods, while different thresholds ($thres$) are investigated for the transition probability of the random walk, used to exclude some unrelated cells, in other words, when $p < thres$, the corresponding cell  is dropped. The results demonstrate that the random walk strategy effectively suppresses noise during context aggregation, thereby enhancing model's performance. Among the strategies, the model performs best with a threshold of 0.67, achieving R, P, and F1 scores of 47.96, 41.32, and 44.39, respectively.}

\begin{table}[htp]
	\centering
	\caption{Comparison with state-of-the-art methods on the NUS-WIDE dataset, where numbers in \textcolor{red}{red} indicate the best performance and numbers in \textcolor{blue}{blue} represent the second-best performance.}
	\label{tab5}
	\resizebox{0.9\linewidth}{!}{
	\begin{tabular}{c|c|c|c|c|c}
		\toprule
		\makebox[0.3\textwidth][c] {\textbf{Method}} & 	\makebox[0.2\textwidth][c] {\textbf{Backbone}} & \makebox[0.15\textwidth][c] {\textbf{cells}} &\makebox[0.1\textwidth][c] {\textbf{mAP}} & \makebox[0.1\textwidth][c] {\textbf{CF1}} & \makebox[0.1\textwidth][c] {\textbf{OF1}}\\
		\midrule
		MS-CMA \cite{you2020cross}     & ResNet101&&61.4&60.5&73.8  \\
		SRN   \cite{zhu2017learning}   &ResNet101        &  & 62.0 & 58.5 & 73.4 \\
		ICME \cite{chen2019multi}      &ResNet101        & &62.8 & 60.7 & 74.1 \\
		ASL  \cite{ridnik2021asymmetric} &ResNet101        &  & 65.2 & 63.6 & \textcolor{red}{75.0}\\
		Q2L-R101   &ResNet101     &    &65.0&63.1 &\textcolor{red}{75.0} \\ 
		ML-SGM   \cite{wu2023semantic}&ResNet101 &  &64.6&	62.4 &	72.5 \\
		SST       \cite{chen2022sst}  &ResNet101& & 63.5 &  59.6 & 73.2 \\
		SADCL \cite{ma2023semantic} & ResNet101 & &\textcolor{blue}{65.9}& \textcolor{blue}{63.0}&\textcolor{red}{75.0} \\
		\multirow{2}{*}{DMCKN (ours)} &ResNet101      & 4*5  &65.4& 63.9&74.2 \\
		&ResNet101      & 8*10  &\textcolor{red}{66.3}&\textcolor{red}{64.6}&\textcolor{blue} {74.8}
		\\ \hline
		Focal loss \cite{lin2017focal} &TresNetL && 64.0 &62.9 &74.7 \\
		ASL     &TresNetL && 65.2 &63.6 &75.0 \\
		Q2L-TResL & TresNetL & &\textcolor{blue}{66.3} &\textcolor{blue}{64.0} &\textcolor{blue}{75.0} \\
		\multirow{2}{*}{DMCkN (ours)} &TresNetL&4*5  &66.9 & 64.5&75.8 \\
		&TresNetL&8*10&  \textcolor{red}{67.8} &\textcolor{red}{65.1}& \textcolor{red}{76.5}\\ 
		\hline
		MlTr-l   &MlTr-l(22k)        &   & 66.3 &65.0 &75.8 \\
		Q2L-CvT \cite{liu2021query2label} &CvT-w24       &   & \textcolor{red}{70.1} & \textcolor{blue}{67.6} & \textcolor{blue}{76.3} \\ 
		\multirow{2}{*}{DMCKN (ours)}&CvT-w24&4*5&69.4&68.2&76.1 \\
		&CvT-w24&8*10  &\textcolor{blue}{69.7}&\textcolor{red}{68.9}& \textcolor{red}{76.6}\\ \bottomrule		
	\end{tabular}
}
	\vspace{-12pt}
\end{table}

\subsection{Results on NUS-WIDE}
The NUS-WIDE dataset is a widely used benchmark for multi-label image classification, comprising 269,648 Flickr images with 5,018 labels and manually annotated with 81 specific concepts,  on average 2.4 concepts per image. According to the official division, 161,789 images are used for training and 107,859 for testing, with small-size images selected for our experiments.

To assess the model's performance on the NUS-WIDE dataset, we employed metrics such as mean Average Precision (mAP), Composite F1 Score (CF1), and Overall F1 Score (OF1), where higher scores stand for better performance. We resized all the images to $400\times500$ pixels and segmented context structures based on 4$\times$5 and 8$\times$10 cell grids.  For the feature extraction, the pre-trained Resnet101, Tresnet, and Cvt models on the ImageNet dataset are used.

Tab.~\ref{tab5} shows  the quantitative results of the proposed method compared to other state-of-the-art on the NUS-WIDE dataset, showing superior performance. With Resnet101, ours achieved improvements of 0.4 and 1.6 in mean Average Precision (mAP) and Composite F1 Score (CF1), respectively. With TresnetL, our model obtained gains of 1.6, 1.1, and 1.5 in mAP, CF1, and Overall F1 Score (OF1), respectively.  Further exploration of  the potential of our method --- employing the latest backbone network (i.e.~CvT-w24) --- allows us to reach  extra  gains of  1.3 in CF1 and 0.3 in OF1.

\begin{figure}[tb]
	\centering
	\includegraphics[width=0.9\linewidth]{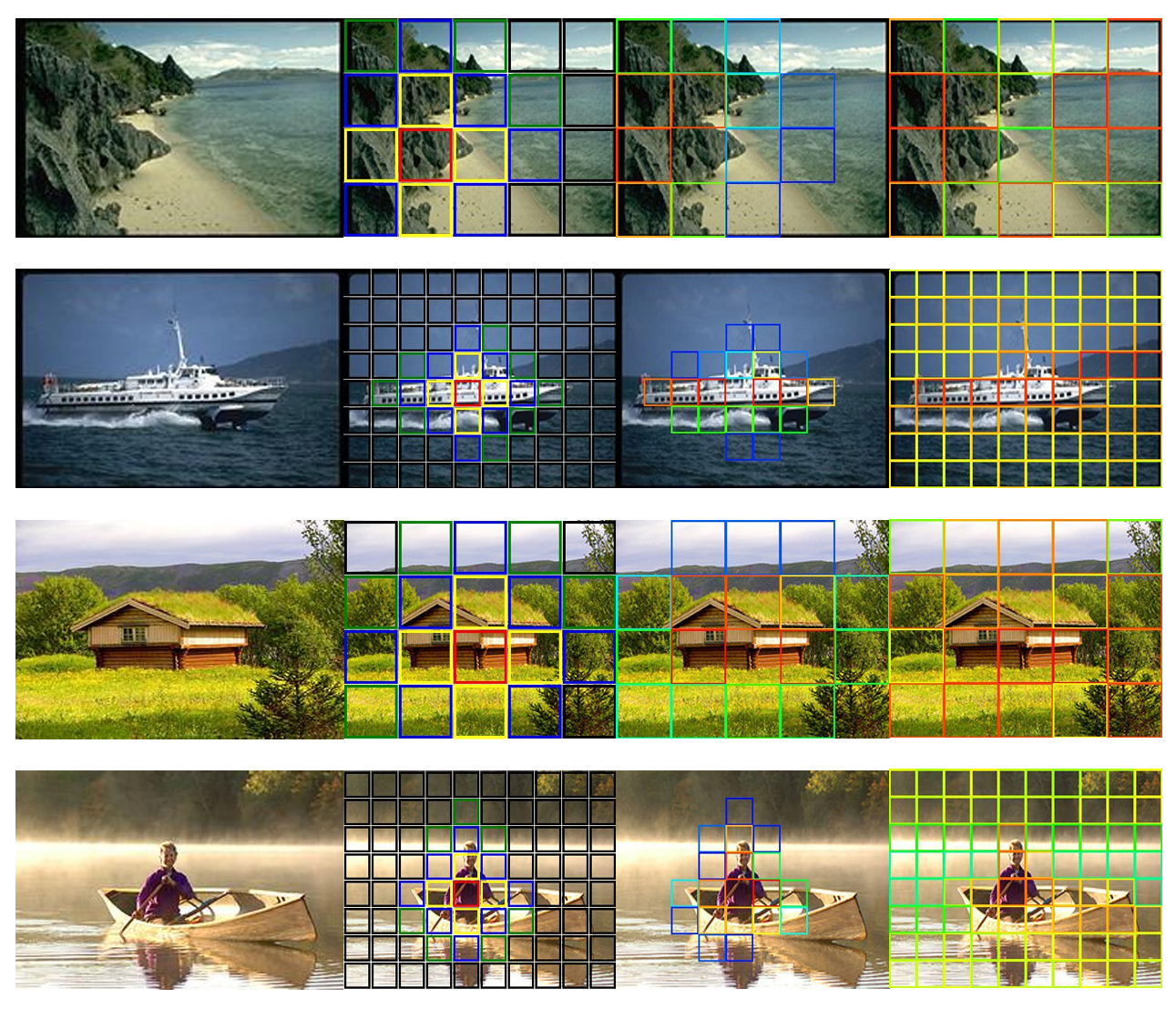} 
	\vspace{-10pt}
	\caption{\small Image instances of the initial and learned context of higher-order domains on the Corel5K dataset (upper half) and the NUS-WIDE dataset (lower half). From the left to right column: the original images, the initial multi-order neighborhood system, the learned different levels of neighborhoods on the central cell, the impacts of different cells. Warmer color stands for higher impact.}
	\label{fig:3}
\end{figure}
\begin{figure}[ht]
	\centering
	\includegraphics[width=0.9\linewidth]{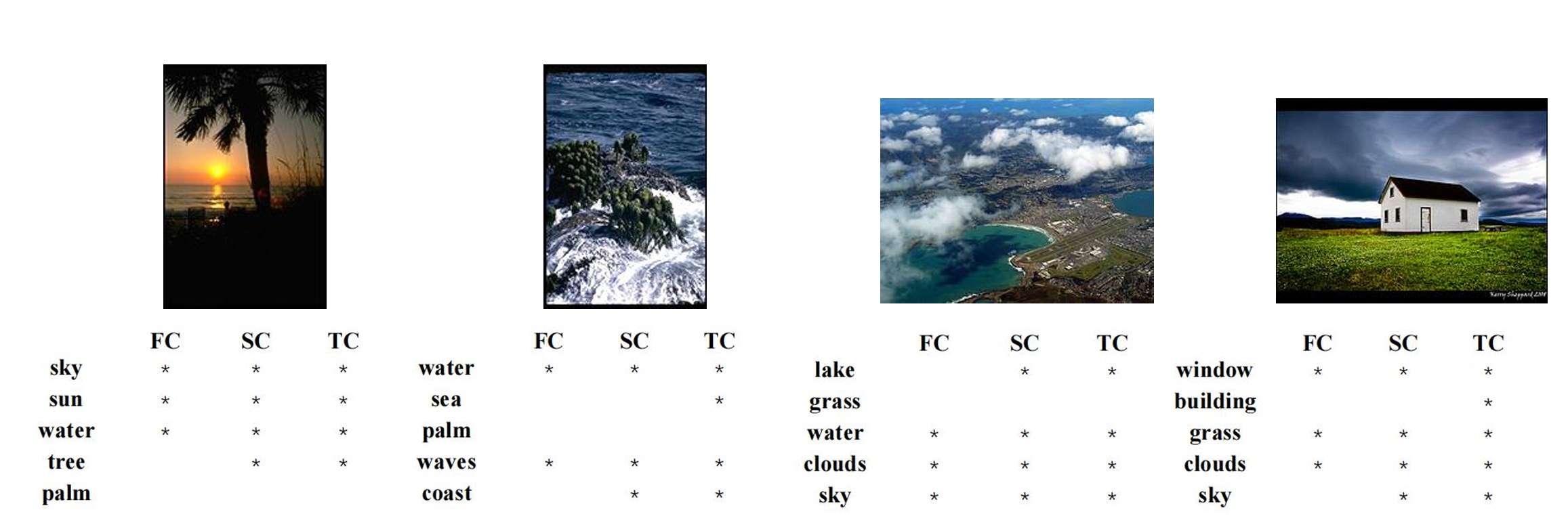} 
	\vspace{-10pt}
	\caption{\small Comparison of image instances of predicted labels and actual labels including FC (First-Order Context), SC (Second-Order Context), and TC (Third-Order Context), the left two images are from the Corel5K dataset and the right two images are from the NUS-WIDE dataset.}
	\label{fig:4}
\end{figure}

\subsection{Visualization of Context Impact and Label Prediction}
Fig.~\ref{fig:3} visualizes the learning effects of our context-aware kernel network on the Corel5K dataset (upper half) and the NUS-WIDE dataset (lower half). The learned multi-order neighbor relationships enhance the focus on visually similar neighboring cells, being capable of capturing more specific and rich contextual information. The final column demonstrates that the network gives higher attention to cells containing both prominent and smaller targets.

Fig.~\ref{fig:4} shows the prediction results of our network. By learning a multi-order neighborhood system, we more precisely identify the detailed features of targets and effectively capture the overall features of images through the integration of contextual information at various levels, significantly enhancing the accuracy of label predictions.

\section{Conclusion}
In this work, we introduce a deep multi-order context-aware kernel network to enhance the multi-label image classification task. By leveraging deep contextual modeling, this approach captures intrinsic structural relationships and external connections, significantly improving classification performance. Our framework aggregates multi-order contextual information, providing more refined feature representations for multi-label learning. Experimental results on the Corel5K and NUS-WIDE datasets validate the effectiveness of our method. Future work will focus on modeling label dependencies within our framework and exploring multi-scale approaches for global image representation. We plan to iteratively merge cells through the context-aware kernel network, which is expected to further boost performance.

\end{document}